\documentclass[runningheads]{llncs}
\usepackage{graphicx}
\usepackage{comment}
\usepackage{multirow}

\usepackage{color}
\usepackage{ulem}
\usepackage{array}

\begin{document}
\title{
Cell Detection from Imperfect Annotation\\ by Pseudo Label Selection Using P-classification
}
\titlerunning{Cell Detection From Imperfect Annotation}
%
\author{Kazuma Fujii\inst{1} \and
Daiki Suehiro\inst{1, 2} \and
Kazuya Nishimura\inst{1} \and
Ryoma Bise\inst{1}}
%
\authorrunning{K. Fujii et al.}
%
\institute{
Kyushu University, Fukuoka, Japan \\
\email{fujii.kazuma@humna.ait.kyushu-u.ac.jp} \\
\email{bise@ait.kyushu-u.ac.jp} \and
RIKEN, AIP, Tokyo, Japan}

\maketitle
\begin{abstract}
Cell detection is an essential task in cell image analysis.
Recent deep learning-based detection methods have achieved very promising results.
In general, these methods require exhaustively annotating the cells in an entire image.
If some of the cells are not annotated (imperfect annotation), the detection performance significantly degrades due to noisy labels. This often occurs in real collaborations with biologists and even in public data-sets.
Our proposed method takes a pseudo labeling approach for cell detection from imperfect annotated data. 
A detection convolutional neural network (CNN) trained using such missing labeled data often produces over-detection. 
We treat partially labeled cells as positive samples and the detected positions except for the labeled cell as unlabeled samples.
Then we select reliable pseudo labels from unlabeled data using recent machine learning techniques; positive-and-unlabeled (PU) learning and P-classification.
Experiments using microscopy images for five different conditions demonstrate the effectiveness of the proposed method.
Our code is available at https://github.com/FujiiKazuma/CDFIAPLSUP.git
\end{abstract}

\section{Introduction}
Cell detection is an essential task in cell image analysis, which has been widely used for cell counting and cell tracking.
Many image processing based methods have been proposed for automatically detecting cells {\it e.g.}, using a thresholding~\cite{otsu1979threshold,yuan2012quantitative,bise2015}, region growing~\cite{zhou2004segmentation} and graph cuts~\cite{al2009improved}. These methods are usually designed on the basis of image characteristics, so they may only work under certain conditions.

Recently, deep learning-based detection methods have shown to be effective for various types of cells if they are trained on enough training data for specific conditions \cite{yi2019multi,li2019signet,fujita2020cell,nishimura2019weakly}.
Moreover, deep learning-based methods usually require fully annotating all the cells in an entire image for the network to learn both the foreground (cell) and the background area. If some of the cells are not annotated ({\it i.e.}, imperfect annotation), the non-annotated cell regions are mistakenly treated as background regions in training. The detection performance significantly degrades due to such noisy labels.
However, it is very costly to annotate all the cells in an image since there are as many as hundreds or thousands of cells in an image. Therefore, some of the current public data-sets only provide partially annotated cells (imperfect annotation).

The aim of our study is to make cell detection feasible from imperfect annotated data by using pseudo labeling.
This would enable the use of imperfect datasets, which have already been made public, and reduce the annotation cost for biologists in real applications.
In the proposed method, we first create a mask that covers only the annotated cells, in which the loss is ignored outside the mask. 
The masked loss facilitates the reduction of false negatives but produces many false positives since it learns part of the foreground (cell) region but not the background. 
When the detector is applied to the dataset that contains partially labeled cells (positive data), the detected positions excluding the labeled cell can be considered unlabeled data, which may contain both positive (cell) and negative (background) positions.
To address the over-detection problem, we propose a semi-supervised method that selects reliable background as pseudo negative labels and additional foreground regions as pseudo positive labels from the unlabeled data and adds these to the training data in the next step.
We applied positive and unlabeled (PU) learning \cite{Kiryo2017PU} to extract the optimal image features that separate the feature distribution of positive (cell) and negative label (background).
In order to minimize the risk of selecting incorrect labels as the pseudo labels, we performed ranking learning using P-classification \cite{Ertekin2011Pclass} which aims to learn a ranking (scoring) function so that reliable positive samples are ranked higher.
These processes are iteratively performed. 
This improves the performance of the detector network by adding reliable labels for both negative and positive positions.
The experiments using microscopy images for five different conditions demonstrate the effectiveness of the proposed method. 

\section{Related work}
\noindent
\textbf{Object detection from imperfect annotation} is more challenging than supervised object detection. Some methods have been proposed for this task in general object detection. Xu {\it et al.} \cite{Xu2019Missing} investigated the effects of imperfect annotation by changing the annotation rate using PASCAL VOC2007. They found that the performance of the current methods such as YOLO \cite{yolov3} and Faster-RCNN \cite{Ren2015FasterRCNN}, drastically degraded as the rate lowered ({\it e.g.}, from 0.7 to 0.45 when the rate was 0.2).
Pseudo labeling has been often used for object detection in semi-supervised learning \cite{kikkawa2019semi,li2019signet,NIPS2019_gberta}, and it could be used for learning from imperfect annotation.
Misra {\it et al.} \cite{Misra2015SSL} proposed semi-supervised learning for object detection from imperfect annotation. In their study, the target object was cars in a video. 
This method selects the reliable positive objects (cars) by tracking candidate objects in the inputted video. However, they used random images from the internet as negative labels, which cannot be used for cell detection tasks.
Recently, Yang {\it et al.} \cite{Yang2020PU} proposed a PU-learning \cite{Kiryo2017PU} based method for general object detection from imperfect annotation. In this method, the positive objects are directly selected using PU learning without re-training the detector. PU learning requires the prior (the proportion of positive samples in the data), and the method assumes a small number of the same class objects in an image. However, in our target, there are many cells in an image, so it is difficult to estimate the prior, in our preliminary work, PU learning was not effective. 

Unlike the previous methods, our method incrementally improves the detector by selecting reliable pseudo positive and negative labels using recent machine learning techniques.

\begin{figure}[t]
    \centering
	\includegraphics[width=\textwidth]{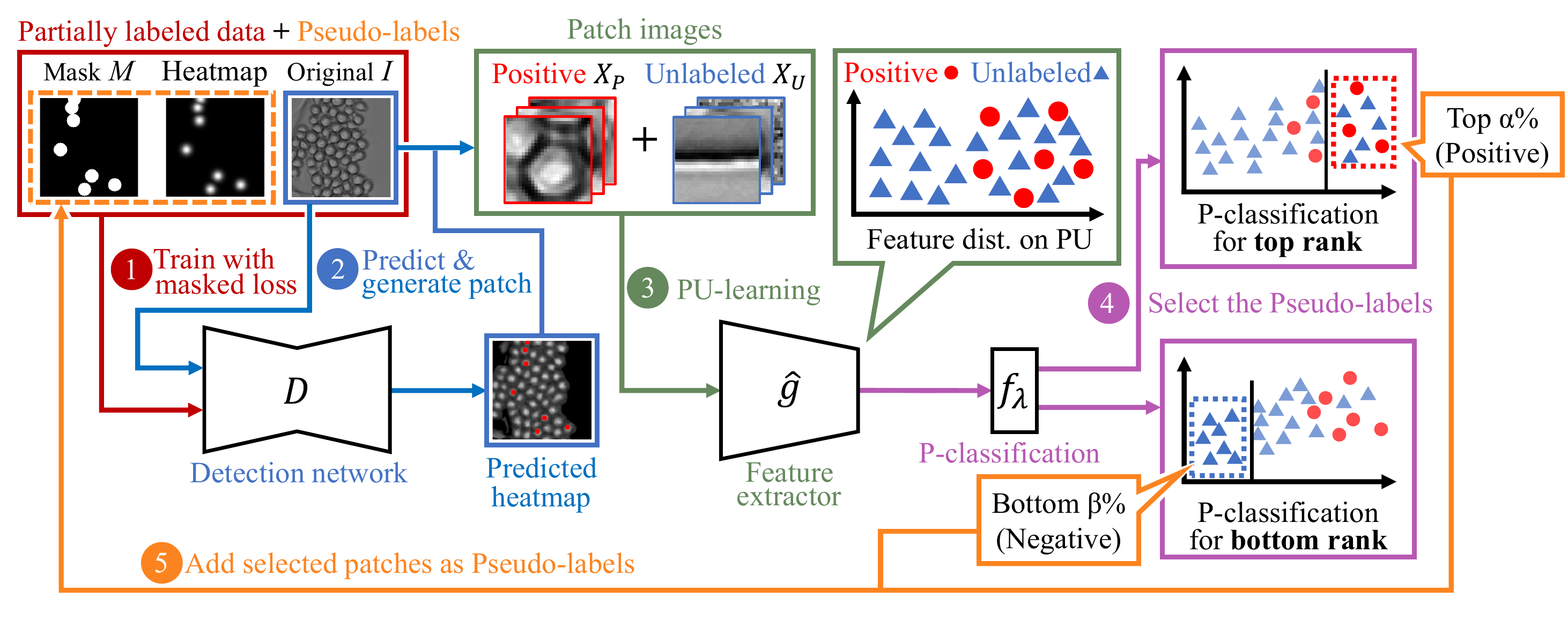}
	\caption{Overview of the proposed method. Given the partially labeled data, to improve the detection network $D$, the proposed method iteratively adds the pseudo labels for both positives and negatives by iteratively performing the five steps; 1) training $D$ with the masked loss using partially labeled data; 2) predicting the position heatmaps and generates patch images from detected points; 3) training the feature extractor $\hat{g}$ using PU learning; 4) training the ranking function $f_{\lambda}$ and selecting pseudo labels for positive and negative samples based on the ranking score; 5) adding the selected patches as pseudo labels to be used for the next iteration.
	}
  \label{fig:overview}
\end{figure}

\section{Cell detection from imperfect annotation}
Fig. \ref{fig:overview} shows an overview of the proposed method.
Given the original images $I$ with the imperfect annotation of the cell position $C$, the method iteratively trains the detection network $D$ that estimates the position heatmap\cite{nishimura2019weakly} by pseudo labeling. In the estimated heatmap, a cell centroid becomes a peak with a Gaussian distribution as shown in the estimated image in Fig. \ref{fig:overview}. The method consists of the following five steps: 
1) training the detection network $D$ using the imperfect annotation for cell centroid positions $C$; 
2) generating patch images based on the detection results; 
3) training the feature extractors using PU-learning \cite{Kiryo2017PU} with positive and unlabeled patch images; 
4) training the ranking function that estimates the score that a sample belongs to positive or negative using P-classification \cite{Ertekin2011Pclass}; 
5) selecting reliable pseudo-labels using the estimated scores, and adding the center positions of the selected patches as the pseudo-labels $C_P$ to be used for training in the next iteration. These steps are iteratively performed and these improve the detection performance of $D$ using the reliable pseudo labels for both positive (cell) and negative (background) positions.

\subsubsection{Cell detection with masked loss and patch generation:}
We use a cell position heatmap that has shown a good performance \cite{nishimura2019weakly}. The detection network $D$ is trained using the set of the annotated cell centroid positions to predict a heatmap so that a cell centroid becomes a peak with a Gaussian distribution. We use the U-net architecture for $D$.
The method requires all cells in an image to be annotated as training data since the network is trained to produce zero values on the background area (non-annotated area).
If we train the network using the imperfect training data, it affects the detection performance.
Therefore, we use the masked loss function which gives the loss only around the given ground-truth of the cell centroids. 
A mask $M_{k}$ is generated for the $k$-th image so that a circular area within radius $r$ from an annotated position becomes the foreground (loss is calculated) and otherwise background. 
The radius of the mask should be large enough to cover a Gaussian distribution in the heat-map. 
Examples of the heatmap and mask are shown in the training data in Fig. \ref{fig:overview}.

Next, the trained model predicts the heatmap for all the original images $I$.
Local maximum points that are larger than a threshold $th$ in the heatmap are detected as cell positions.
The masked loss facilitates the reduction of the false negatives but produces many false positives. 
We have to select the reliable detection results from the noisy prediction to effectively perform pseudo labeling.
The patch images are first generated by cropping the cell image based on the detected points. 
The patch image with the annotated positions ${C}$ can be considered positive labeled samples ${X}_P=\{x_i^p\}_{i=1,...,N_p}$, and the patch images except ${X}_P$ can be considered the unlabeled samples ${X}_U=\{x_i^u\}_{i=1,...,N_u}$.

\subsubsection{Feature extraction using PU-leaning:}
The aim of the 3rd step is to extract optimal image features that separate the feature distribution of positive (cell) and negative label (background), which is useful to select reliable labels. 
To achieve this, we use PU learning \cite{Kiryo2017PU} which uses positive labeled samples and unlabeled samples as training data, so the unlabeled samples can be classified as positive or negative.

Let $g$ be the classifier CNN that classifies an input patch image $x_i$ into positive (+1:cell) or negative (-1:background), where training is performed using positive and unlabeled data.
In PU-learning, the following non-negative risk function \cite{Kiryo2017PU} is used as the CNN loss:
\begin{equation}\label{eq:eachlikely}
     {R}_{pu}(g) = \pi_{p} {R}_{p}^+(g) + \max \{ 0, {R}_{u}^-(g) - \pi_{p} {R}_{p}^-(g) \},
\end{equation}
where $\pi_{p}$ represents $prior$ probability of the positive samples ({\it i.e.}, the proportion of the positive samples in the data), ${R}_{p}^+(g)=(1/N_p)\sum_{i=1}^{N_p}{l(g(x_i^p),+1)}$, ${R}_{p}^-(g)=(1/N_p)\sum_{i=1}^{N_p}{l(g(x_i^p),-1)}$ and ${R}_{u}^-(g)=(1/N_u)\sum_{i=1}^{N_u}{l(g(x_i^u),-1)}$. $l$ is by default the {\it zero-one} loss, namely $l_{01}(t,y)=(1-\mathrm{sign}(ty))/2$.

After training, we obtain the feature extraction layers $\hat{g}$ by removing the FC classifier layer from $g$. 
We use the image features $\hat{g}(x_i)$ in the next step.

\subsubsection{Pseudo-label selection using P-classification:}
Although PU learning aims to extract optimal features for distinguishing positive and negative samples, the proportion is unknown in real application and the distribution of the features depends on the dataset.
Therefore, it is difficult to directly determine the discriminant plane in the feature space, and it may contain many incorrect labels.
Instead, we take the pseudo labeling approach to select reliable samples from the unlabeled data $X_U$ based on the extracted features.

To minimize the risk of selecting incorrect labels as pseudo labels, we use P-classification \cite{Ertekin2011Pclass} which is a technique for learning to rank. 
The advantage of using P-classification is that we can emphasize the classification performance at the top of the ranked list.
More precisely, P-classification aims to learn a ranking (scoring) function which gives large scores to (possibly a limited number of) reliable positive samples rather than gives large scores for all positive samples. 
This indicates that we can reduce the risk that the high rank scored samples contain negative (incorrect) samples, ({\it i.e.}, the precision on the higher-ranked samples becomes high).

Given the positive $\{x_i^p\}_{i=1}^{N_p}$ and negative samples $\{x_k^n\}_{k=1}^{N_n}$, P-classification optimizes the parameters of a scoring linear function $f_{\lambda}:=\sum_j{\lambda_j g_j}$, in which $g_j$ is the $j$-th feature of $\hat{g}(X_i)$, using the following loss:
\begin{equation}\label{eq:eachlikely2}
     {\cal R}^{PC}( \mathbf{\lambda} ) := \sum^{N_p}_{i=1} e ^{ -f_{\mathbf{\lambda}} (x_{i}^p) } + \frac{1}{p} \sum^{N_n}_{k=1} e^{ f_{\mathbf{\lambda}} (x_{k}^n) }, 
\end{equation}
where $p~ (\geq 1)$ is a hyper-parameter which controls the degree of concentration at the top of the list. Roughly speaking, when we set $p=1$, the method maximizes AUC (Area Under the receiver operating characteristic Curve), which is a standard scoring/ranking performance measure. When we set a larger $p$, the method maximizes the leftmost portion of the curve ({\it i.e.}, performance at the top of the ranked list).
This method only trains the discriminant layer, and thus we used $\hat{g}$ as the feature extractor network.  

In this study, the negative samples are unknown and the objective is to reduce the risk of selecting incorrect labels as the pseudo-labels. Therefore, we use the unlabeled data $X_U$ as $\{x_k^n\}_{k=1,...,N_n}$, instead. 
Then we select the top $\alpha$ of the ranked unlabeled data is selected as the positive pseudo-labels.
In addition, we also have to select the negative samples (background) from $X_U$ to reduce the false positives. To select bottom-negatives, we use P-classification by opposing the labels. Then, the bottom $\beta$ of the ranked unlabeled data is selected as the negative pseudo-labels.

From the selected pseudo-labels, we re-generate the pseudo ground-truth of the position heatmap and the mask that will be used in the next iteration.
The center positions of the selected positive pseudo patches are additionally registered as pseudo labels $C_P$. It is used for both generating pseudo position heatmap and the mask for the training loss. The center positions of the selected negative pseudo patches are used for generating the mask, where a position in the pseudo heatmap has a value of 0 in the background. These steps are iteratively performed and improve the detection performance of $D$ using the reliable pseudo labels for both positive (cell) and negative (background) positions.

\section{Experiment}
\begin{figure}[t]
    \centering
	\includegraphics[width=0.95\textwidth]{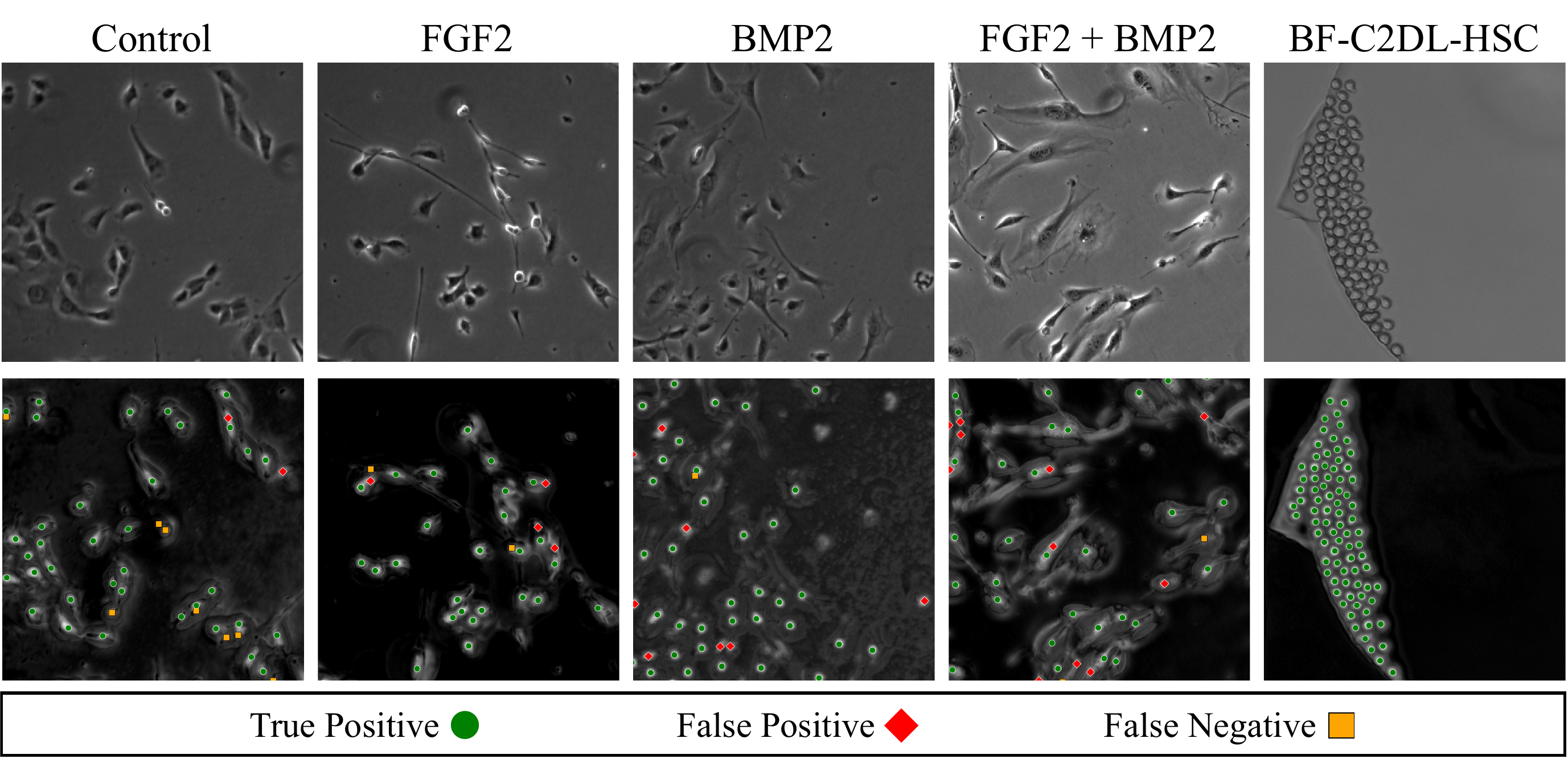}
	\caption{Examples of enlarged images of detection results in each condition.}
  \label{fig:experiment_result}
\end{figure}

\subsubsection{Dataset and performance metric:}
In the experiments, we used the five conditioned data from two datasets, BF-C2DL-HSC (HSC) from the Cell Tracking Challenge at ISBI \cite{redmon2016you,ulman2017objective} and C2C12 \cite{eom2018phase}.
HSC has a resolution of $1010 \times 1010$ pixels.
Since we assume that a large number of cells exist in a single image and only some of them can be annotated, we used frames with a large number of cells for our experiments.
We used 264 images containing over 80 cells. Thirteen images with partially labeled cells were used for training, in which the total number of annotated cells is 125. The other unlabeled cells in the partially labeled images were used for unlabeled data.
We used the other 251 images for test. In this data, noises appeared on the well, and many false positives were detected in the initial estimation.
In C2C12 \cite{eom2018phase}, myoblast cells were captured by phase contrast microscopy at a resolution of $1040 \times 1392$ pixels and cultured on four different conditions: Control (no growth factor), where 10 images with partially labeled cells and 90 images are used as training and test data, respectively; FGF2 (fibroblast growth factor), with 38 images for training and 342 images for testing; BMP2 (bone morphogenetic protein), with 10 images for training and 90 images for test, 2-4) FGF2+BMP2, with 10 images for training and 90 images for test. 
These cells have more varied cell appearances compared with HSC.
In detection, the heatmap was normalized from $0$ to $255$ and the threshold $th$ was set to $128$, and Mean squared error was used as loss function.
The proportions of the labeled cells in a training image were 0.1 and the number of iterations was 2 in all experiments.
The percentages of pseudo labels to be selected, $\alpha$ and $\beta$, were set to 5 \% in all our experiments, in which we did not perform optimal parameter search for a fair comparison. 
These parameters work well as long as it is not set to an extremely large value.
For $p$ in P-classification, we used 4, which was used in the reference paper\cite{Ertekin2011Pclass}. This $p$-value will work if it is not too small or too large.

We used F-score as the detection performance metric. To compute F-score, we first assigned the positions in detection and ground-truth by one-by-one matching with a certain distance threshold, defined by the cell radius. We used 15 pixels for all data. Then we defined a detection position successfully assigned to a ground-truth as true positives, un-associated detected positions as false positives, and un-associated ground-truths as false negatives.

\begin{table}[t]
\begin{center}
\caption{Performance of the compared methods on each dataset.}
\label{tab1:comparison}
\begin{tabular}{lc|cccccc}
\hline
Data & 
Condition & 
\begin{tabular}{c}Nishimura\\ \cite{nishimura2019weakly}\end{tabular}& 
\begin{tabular}{c}With \\ mask\end{tabular} &
\begin{tabular}{c}Vicar \\ \cite{vicar2019cell}\end{tabular} &
\begin{tabular}{c}Yang \\ \cite{Yang2020PU}\end{tabular} & 
\begin{tabular}{c}Ours \end{tabular} &
\begin{tabular}{c}Fully \\ supervised\cite{nishimura2019weakly}\end{tabular} \\
\hline\hline
\multirow{4}{*}{C2C12} & Control    & 0.477 & 0.239 & 0.802 & 0.239             &\textbf{0.834} & 0.922 \\
                       & FGF2       & 0.318 & 0.361 & 0.648 & \textbf{0.768}    &\textbf{0.768} & 0.924 \\
                       & BMP2       & 0.609 & 0.217 & 0.739 & \textbf{0.809}    & 0.769         & 0.978 \\
                       & FGF2 + BMP2& 0.105 & 0.257 & 0.539 & 0.257             &\textbf{0.578} & 0.962 \\
\hline
\multirow{1}{*}{HSC} &              & 0.157 & 0.393 & 0.475 & 0.108             &\textbf{0.952} & 0.998 \\
\hline\hline
\multirow{1}{*}{} & Average         & 0.333 & 0.293 & 0.6406 & 0.436            &\textbf{0.780} & 0.971 \\
\hline\hline
\end{tabular}
\end{center}
\end{table}

\subsubsection{Evaluation:}
We evaluated the performance of our method using five data with three methods as ablation study; 1) Nishimura \cite{nishimura2019weakly} that was a supervised detection method, it was simply trained using missing labels; 
2) With mask, in which the detector \cite{nishimura2019weakly} was trained with the masked loss; 
3) Vicar \cite{vicar2019cell} that is an image processing-based method using Yin \cite{yin2012understanding} and distance transform \cite{thirusittampalam2013novel};
4) Yang \cite{Yang2020PU} that uses PU-learning to directly determine the labels without pseudo labeling, in which the class prior was estimated by \cite{Plessis2016Prior}.
The idea was simply applied to our model, 
Tab. \ref{tab1:comparison} shows the F-scores of these methods.
The performance of our method was significantly better than the baseline method (Nishimura).
The baseline method treated a lot of cells as background and thus degraded its performance. 
Supervised detection with masked loss is also worse.
The performance from Vicar that is not a learning method is better than the baseline, however, it depends on the image characteristic and thus the performance was not well in some conditions.
Yang \cite{Yang2020PU} improved the performance in some data-sets.
However, the performances in some data were significantly worse.
In our method, detection results using the detection CNN trained by sparse supervised data is unstable, and thus the prior of the positive and negative samples in the test data is also unstable.
Even if we use the correct prior in training, the prior in the test may be different from that in training. 
In this case, the PU-learning could not work well as the discriminator.
For HSC, results of our method achived accuracy close to that of the fully-supervised.
Our method further improved the performance in most of the data. 
For FGF2+BMP2, the further iteration improved the performance (0.750 at iteration 5).
Fig. \ref{fig:experiment_result} shows the examples of our final detection results in each condition. In the results, there are several false positives (two peaks appear in a cell region) but few false negatives.

\begin{figure}[t]
    \centering
	\includegraphics[width=0.95\textwidth]{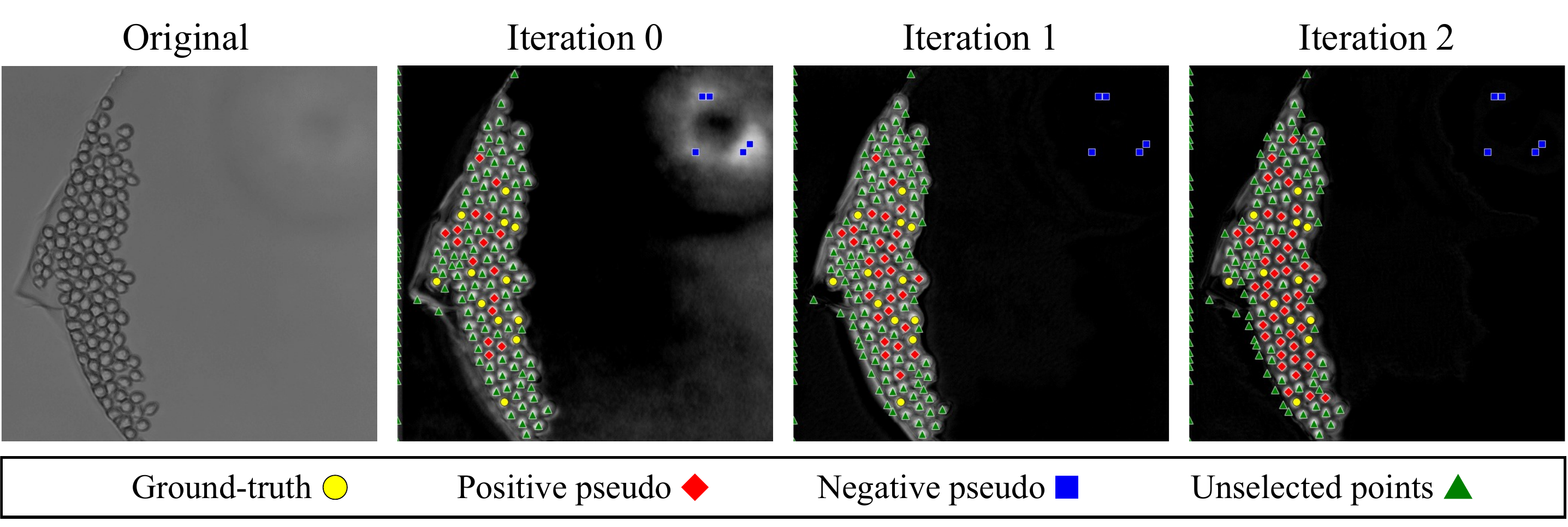}
	\caption{Examples of selected pseudo labels in each iteration.}
  \label{fig:experiment_result2}
\end{figure}

\begin{table}[tbp]
\begin{minipage}{.49\textwidth}
 \caption{Improvements in training.}
 \centering
  \begin{tabular}{cccc}
    \hline
    ittr. & precision & recall & F-score \\
    \hline \hline
    0 & 0.337 & 0.999 & 0.504 \\
    1 & 0.732 & 0.997 & 0.844 \\
    2 & 0.963 & 0.998 & 0.980 \\
    3 & 0.956 & 0.997 & 0.976 \\
    4 & 0.946 & 0.995 & 0.970 \\
    5 & 0.952 & 0.996 & 0.973 \\
    \hline
  \end{tabular}
 \label{tab:evaltrain}
\end{minipage}
\begin{minipage}{.49\textwidth}
 \centering
 \caption{Improvements in test.}
  \begin{tabular}{cccc} 
    \hline
    ittr. & precision & recall & F-score \\ 
    \hline \hline
    0 & 0.245 & 0.999 & 0.394 \\
    1 & 0.622 & 0.999 & 0.767 \\
    2 & 0.910 & 0.999 & 0.952 \\
    3 & 0.896 & 0.998 & 0.944 \\
    4 & 0.918 & 0.997 & 0.956 \\
    5 & 0.888 & 0.998 & 0.940 \\ 
    \hline
  \end{tabular}
 \label{tab:evaltest}
\end{minipage}
\end{table}

\subsubsection{Improvement process in training:}
Fig. \ref{fig:experiment_result2} shows the improvement process in iterations using HSC. In this experiments, the yellow points are given as the initial partial labels (Ground-truth). 
The green points are predicted to be cells by the detection network but not selected as pseudo-labeled points via P-classification.
The image at iteration 0 indicates the prediction results, where the background region also has high values, so there are false positives. In iteration 0, the positive (red) and negative (blue) positions were successfully selected as pseudo labels. In iteration 1, the estimation of the background was improved. After the first iteration, the unselected detection points (green) are also good.
We can observe that the positive pseudo positions (red) increase with the iteration. 

Tables \ref{tab:evaltrain} and \ref{tab:evaltest} show the precision, recall, and F-score in each iteration under train and test data , respectively.
'ittr.' indicates the number of iteration for pseudo labeling.
In the training phase, the recall of the initial estimation is very high but the precision is low. It supports our assumption that the results estimated using the masked loss have many false positives but few false negatives.
We can observe that the precision was improved with keeping high recall by applying our method. Here, the improvements were saturated after the second iteration in both train and test data. This shows that our iterative pseudo learning approach improved the precision and F-score, and the few iterations worked enough.
We consider that the small number of iteration does not affect the performance and improves the performance for many datasets, even though the additional iteration may improve the performance.
It is our future work to find an optimal iteration using the relationship between the feature distribution and the selected pseudo labels.

\section{Conclusion}
We propose a cell detection method that can train the detection network from imperfect annotation, where only partial cells are annotated in an image.
To archive this, our method selects reliable positive and negative samples from the over-detection results using the recent machine learning approaches: positive-and-unlabeled learning and P-classification.
The experiments using microscopy images for five different conditions demonstrated the effectiveness of the proposed method, which can significantly improve the performance using very few labeled cells (only 10 \% of the total cells). 

{\bf Acknowledgment:}
This work was supported by JSPS KAKENHI Grant Number JP19K22895 and JP20H04211.

\bibliographystyle{splncs04}
\bibliography{refer}

\end{document}